\documentclass[runningheads]{llncs}
\usepackage[T1]{fontenc}
\usepackage{graphicx}
\usepackage{amsmath}
\usepackage{algorithmic}
\usepackage{algorithm}
\usepackage{multirow}
\usepackage{booktabs}
\usepackage{arydshln}
\usepackage[table]{xcolor}
\usepackage{graphicx}

\begin{document}
\title{Locating~Tennis~Ball~Impact on~the~Racket in~Real~Time Using~an~Event~Camera}
%
\titlerunning{Locating~Tennis~Ball~Impact Using~an~Event~Camera}
%
\author{Yuto~Kase\orcidID{0009-0002-1512-0331} \and
Kai~Ishibe\orcidID{0009-0004-0470-9540} \and
Ryoma~Yasuda\orcidID{0009-0000-2620-1896} \and
Yudai~Washida\orcidID{0009-0007-3466-7451} \and
Sakiko~Hashimoto\orcidID{0009-0002-9536-5354}
}
\authorrunning{Y. Kase et al.}
%
\institute{Mizuno Corporation, 1-12-35 Nanko Kita, Suminoe-ku, Osaka, Japan \\
\email{\{ykase, kishibe, ryasuda, ywashida, skhashim\}@mizuno.co.jp}\\
\url{https://corp.mizuno.com}}
\maketitle              
\begin{abstract}
In racket sports, such as tennis, locating the ball's position at impact is important in clarifying player and equipment characteristics, thereby aiding in personalized equipment design.
High-speed cameras are used to measure the impact location; however, their excessive memory consumption limits prolonged scene capture, and manual digitization for position detection is time-consuming and prone to human error. 
These limitations make it difficult to effectively capture the entire playing scene, hindering the ability to analyze the player's performance.
We propose a method for locating the tennis ball impact on the racket in real time using an event camera.
Event cameras efficiently measure brightness changes (called `events') with microsecond accuracy under high-speed motion while using lower memory consumption.
These cameras enable users to continuously monitor their performance over extended periods.
Our method consists of three identification steps: time range of swing, timing at impact, and contours of ball and racket.
Conventional computer vision techniques are utilized along with an original event-based processing to detect the timing at impact (PATS: the amount of polarity asymmetry in time symmetry).
The results of the experiments were within the permissible range for measuring tennis players' performance. Moreover, the computation time was sufficiently short for real-time applications. 

\end{abstract}
%
\begin{figure}[t]
  \centering
   \includegraphics[width=1.0\linewidth]{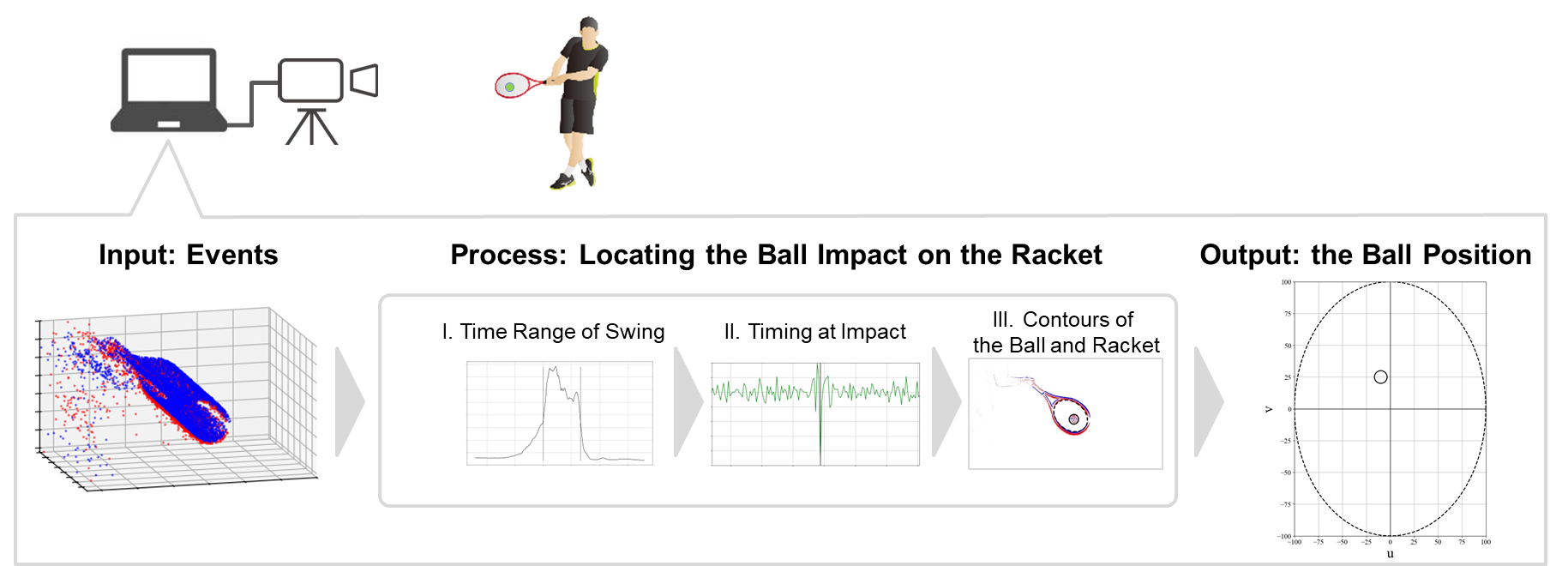}
   \caption{Overview of the proposed method.
   A set of events captured by an event camera while playing tennis are input; $+$ polarity events are represented in blue and $-$ polarity events are represented in red.
   Locating the ball impact on the racket involves three steps.
   The output is the visualization of the ball position, which is relative to the racket.}
   \label{fig:overview}
\end{figure}

\section{Introduction}
\label{sec:intro}
In racket sports, such as tennis, increasing the ball velocity is a key factor. 
Na{\ss} \textit{et~al.\,}\cite{Daniela98} confirmed that the velocity depends on the impact location because the different restitution coefficients of the racket surface vary depending on the area, and the impact locations are also quite different among players. 
Therefore, the measurement of the ball location for each player can aid in personalized equipment design.

To measure the impact location, conventional frame-based cameras (high-speed cameras) or motion capturing systems are utilized, as demonstrated by Ikenaga \textit{et~al.\,}\cite{Masahiro20}.
High-speed cameras can accurately measure positions while preventing motion blur; however, they have a limitation regarding high memory consumption, consuming a large amount of memory as shutter speed increases, which limits the capturing period.
This hinders capturing actual playing scenes, making it difficult to analyze players' performance efficiently.
Motion capturing systems can accurately measure the marker positions, which need to be pre-attached to the ball and racket.
As stated by Karditsas \cite{karditsas2020large}, to obtain the impact location from the data captured by these systems, manual digitization is a valid process; however, the digitization of multiple points across many images is a time-consuming process that is prone to human error. 
Furthermore, it is noted that automated processing offers an efficient and effective solution to this issue.
To address these problems, we propose a method for locating the tennis ball impact on the racket in real time using an event camera.

Event cameras (also known as event-based cameras or dynamic vision sensors) \cite{CenturyArks25,iniVation25,Prophesee25} differ from conventional frame-based cameras in that instead of capturing full images at a fixed rate, they asynchronously measure per-pixel brightness changes and output a stream of events that encode the time, location, and sign of the brightness changes.
This sign is referred to as polarity with `$\mathrm{+}$' and `$\mathrm{-}$' representing the brightness increase and decrease, respectively.
These cameras offer attractive properties: high temporal resolution (on the order of microseconds) resulting in reduced motion blur, low data volume, low latency, low power consumption, and high dynamic range.
In sports, these technologies enable users to easily measure and analyze their performance and equipment.
In sports research, methods utilizing event cameras have been proposed to interpolate frame rates using a frame-based camera by  Deckyvere \textit{et~al.\,}\cite{Antoine24}, estimate the ball's rotation rate by Gossard \textit{et~al.\,}\cite{Thomas24} and Nakabayashi \textit{et~al.\,}\cite{Takuya24}, and detect the position of the ball's trajectory by Nakabayashi \textit{et~al.\,}\cite{Takuya23}.

Our main contributions are summarized as follows:
\renewcommand{\labelitemi}{$\bullet$}
\begin{itemize}
 \item A method to locate the tennis ball impact on the racket in real time (Fig.\,\ref{fig:overview}).
 \item An original event-based processing method to detect the timing at impact, referred to as the amount of polarity asymmetry in time symmetry (PATS).

\end{itemize}
\section{Related Work}
\label{sec:relatedwork}
\subsection{Impact Location Methods in Sports}
\label{subsec:imapcatLocationMethodInSports}
There are two automated approaches based on sensing systems for locating the ball position upon impact:

The first is an inertial measurement unit (IMU) installed in a tennis racket, as developed by Yamashita and Matsunaga \cite{Kosei13}.
It identifies the ball location at impact using the vibration data resulting from the impact between the racket and the ball.
The data are subjected to frequency analysis, using methods such as the Fast Fourier Transform (FFT), whereby the system matches the frequency characteristics with a database storing the characteristics of each racket to identify the ball location.
This approach has the advantages of a less burdensome setup and prolonged scene capture.
However, it requires the database to estimate the impact area of the racket.

The second requires high-speed cameras to be installed in a stationary device, as developed by Kiraly and Merloti \cite{Chris13} and Kiraly and Wintriss \cite{Chris03}.
They identify the ball location at impact for golf using object detection methods, such as the Hough transform, applied to the image at the moment of impact.
These approaches have the advantage of precisely measuring the ball location.
To detect the timing at impact, an image subtraction method is used, which subtracts the current image from the previous image pixel by pixel, and then compares it to a threshold.
This method reduces the memory consumption for high-speed image storage.
Although useful in golf, where the impact location is fixed, it is less suitable for tennis, where the impact location may vary and the player's body may enter the frame.
Additionally, another system that combines a high-speed camera system with Doppler radars has been developed by Fredrick \cite{Fredrick19} to trigger at impact.

These existing approaches make it difficult to accurately locate the impact of the ball in tennis. 
Therefore, our method aims to overcome this challenge by using an event camera.


\subsection{Event Camera Representation}
\label{subsec:event-basedCameraRepresentation}
In using an event camera, each event includes x, y, polarity, and time retrieved in chronological order.
The events are transformed into various alternative representations to facilitate the extraction of meaningful information and solve a particular task, as shown in the survey \cite{Guillermo22} of Gallego \textit{et~al.\,} 
The following representations are related to our method.

The first is an event packet.
This representation is a set of events within the specified accumulation time interval.
The number of events in the packet per second is called the event rate, as indicated by Gossard \textit{et~al.\,}\cite{Thomas24}.
This representation has the advantages of precise timestamp, the number of events, and low computational complexity.
However, it lacks meaningful information, such as the coordinates and polarity.

The second is an event image.
This representation is an image where each pixel stores a polarity value related to the last event at that pixel, as indicated by Prophesee \cite{PropheseeBaseFrame}.
It has the advantage of being usable with conventional image-based computer vision algorithms.
However, the time information can be lost.

The third is a time surface.
This representation is an image where each pixel stores a single time value related to the last event at that pixel, generated separately for each polarity value, as indicated by Lagorce \textit{et~al.\,}\cite{Lagorce17} and Sironi \textit{et~al.\,}\cite{Sironi18}.
The value of each pixel in the image is higher for more recent events, as in an intensity map.
This has the advantages of being compatible with conventional image-based computer vision algorithms, similar to the event image while retaining the time information.
However, this representation  cannot simultaneously handle both polarities.

\begin{figure}[t]
  \centering
   \includegraphics[width=1.0\linewidth]{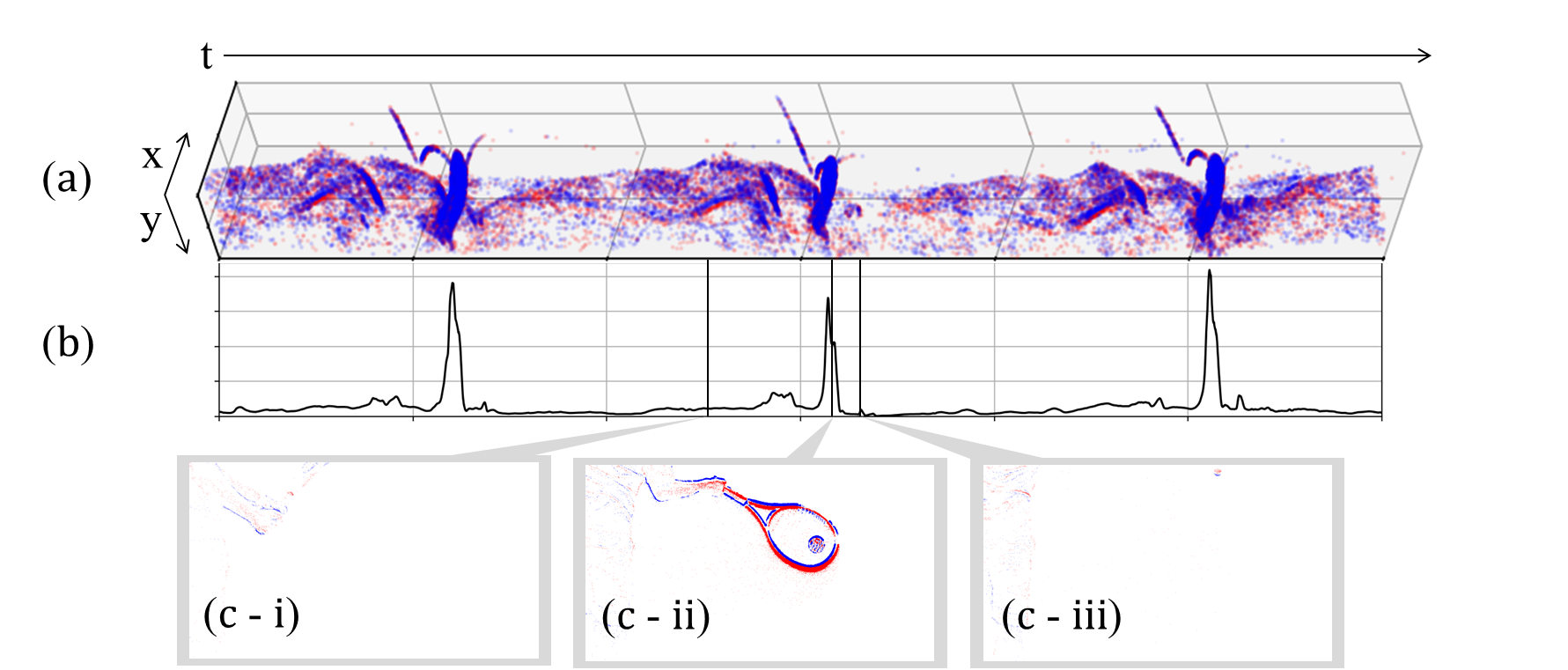}
   \caption{Time series of event packets $\varepsilon_t$ in tennis scenes.
   (a) event data $e_k$ colored according to polarity ($+$ in blue, $-$ in red);
   the areas with smaller x-coordinates show clusters of events caused by racket swings, whereas the areas with larger x-coordinates show clusters of events caused by ball bounces.
   (b) time series of event rates ($| \varepsilon_t |$ per second) which increase while a player is swinging the racket; this player swings three times in these scenes. 
   (c) event image at a specific point in time. Image (c - i) is captured before swinging; image (c - ii) is captured at impact; and image (c - iii) is captured after swinging.}
   \label{fig:time_range_of_swing}
\end{figure}

In this paper, we represent the event packet for input (Fig.\,\ref{fig:time_range_of_swing} (a)), event rate of the event packet to identify the time range of the swing (Fig.\,\ref{fig:time_range_of_swing} (b)), and event image to identify the contours of the ball and racket (Fig.\,\ref{fig:time_range_of_swing} (c)). 
Inspired by the event image and time surface, we propose an original processing method to detect the timing at impact (Fig.\,\ref{fig:timing_at_impact}).

\section{Method}
\label{sec:method}
As shown in Fig.\,\ref{fig:overview}, the proposed method can be divided into three parts: input (Sec.\,\ref{subsec:input}), process (Sec.\,\ref{subsubsec:timeRangeOfSwing}--\ref{subsubsec:contoursOfBallAndRacket}), and output (Sec.\,\ref{subsec:output}).
Each input is a set of events; the process identifies the contours of the ball and racket at impact; and the output is the relative position of the ball on the racket.
The process consists of three identification steps: timing range of swing (Sec.\,\ref{subsubsec:timeRangeOfSwing}), timing at impact (Sec.\,\ref{subsubsec:timingAtImpact}), and contours of ball and racket (Sec.\,\ref{subsubsec:contoursOfBallAndRacket}).

\subsection{Input}
\label{subsec:input}
Individual events captured by a camera are expressed as $e_k = ( x_k, y_k, p_k, t_k)$, where $k$ is an index for each event retrieved in chronological order; $x_k$ and $y_k$ are pixel coordinates $(\mathrm{px})$ with intervals of $[0, width-1]$ and $[0, height-1]$ based on the camera's pixel size; $t_k$ is a timestamp in microseconds ($\mu \mathrm{s}$); and $p_k$ is the polarity ($p_k \in \{\mathrm{+}, \mathrm{-}\}$).

A set of events within the specified accumulation time interval is called an event packet.
The event packet $\varepsilon_t$ is defined as
\begin{equation}
\varepsilon_t = \left\{ {e_k} \middle| {t} {-} {\frac{t_{acc}}{2}} < {t_k} \leq {t} {+} {\frac{t_{acc}}{2}} \right\} ,
\label{eq:eventPacket}
\end{equation}
where $t$ is the reference time $(\mu \mathrm{s})$, and $t_{acc}$ is the accumulation time $(\mu \mathrm{s})$.
Therefore, $\varepsilon_t$ is a $| \varepsilon_t | \times 4$ matrix, which represents the attributes $( x_k, y_k, p_k, t_k)$ as columns, and the event packet $\varepsilon_t$ is the input.

\subsection{Time Range of Swing}
\label{subsubsec:timeRangeOfSwing}
As shown in Fig.\,\ref{fig:time_range_of_swing}, when a player starts to swing the racket, the number of events in the packet $| \varepsilon_t |$ increases.
The racket moves relatively faster than other objects, including the ball, player's body, and background.
To determine the swing, we utilize $| \varepsilon_t |$ per second, called event rates, as explained in Sec.\,\ref{subsec:event-basedCameraRepresentation}.
A time series of event rates identify the time range of the swing as follows.

As the reference time $t$ progresses at constant intervals with stride time $t_{strd}$, the event rate $| \varepsilon_t |$ is calculated for each packet.
The mean and variance for the consecutive $n_{\varepsilon}$ event rates are then determined.

\begin{figure}[t]
  \centering
   \includegraphics[width=0.6\linewidth]{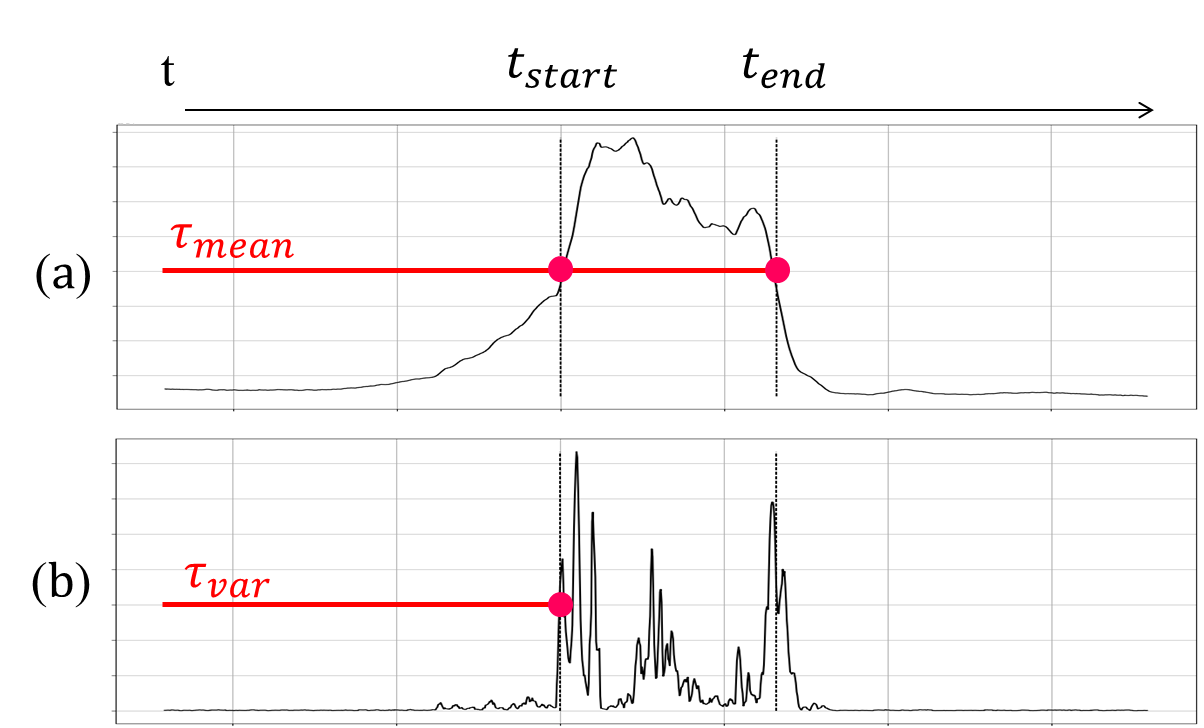}
   \caption{Time series of the mean and variance for the consecutive $n_{\varepsilon}$ event rates.
   (a)~time series of the mean, where the red solid line represents the threshold $\tau_{mean}$.
   (b)~time series of the variance, where the solid red line represents the threshold $\tau_{var}$.
   The black dashed lines represent $t_{start}$ and $t_{end}$, respectively.
   $\tau_{mean}$ and $\tau_{var}$ are used to identify $t_{start}$, while $\tau_{mean}$ is used for $t_{end}$.
   }
   \label{fig:time_series_event_rates}
\end{figure}

The start time of the swing $t_{start}$ is defined as the first moment that exceeds both the mean threshold $\tau_{mean}$ and the variance threshold $\tau_{var}$.
The end time of the swing $t_{end}$ is defined as the first moment that occurs after a time interval $\tau_{t}$ and falls below $\tau_{mean}$, as shown in Fig.\,\ref{fig:time_series_event_rates}.

Subsequently, we detect the time at impact $t_{imp}$ in the time range of the swing $[t_{start}$, $t_{end}]$.

\begin{figure}[t]
  \centering
   \includegraphics[width=0.6\linewidth]{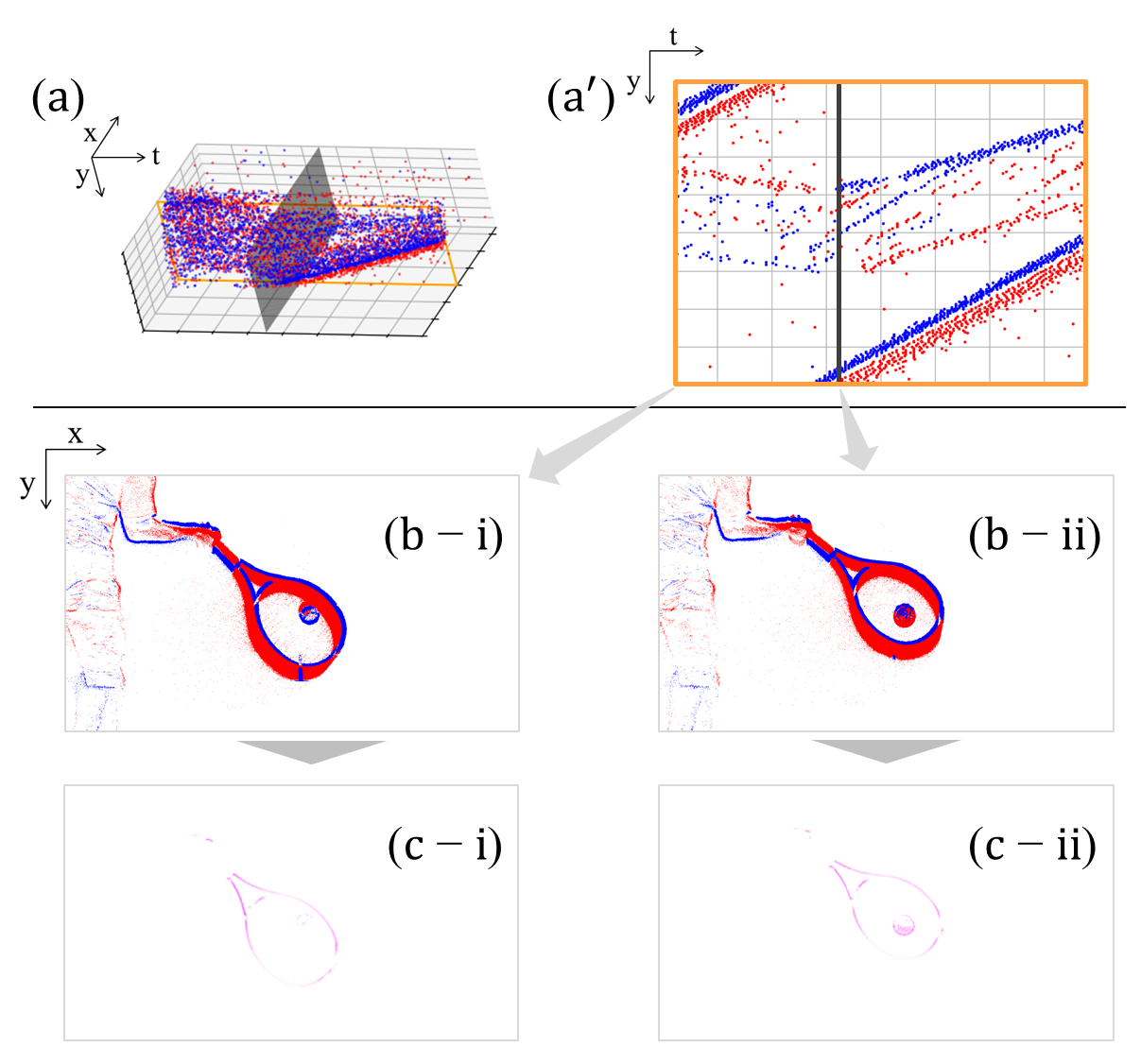}
   \caption{Comparison of the visualization of PATS images before and at impact.
   (a) 3D plot of events $e_k$ near the impact timing $t_{imp}$, where the blue points represent $+$; the red points represent $-$; the black plane represents $t_{imp}$; and the orange edge represents a plane vertical to the t-axis that contains the ball events. 
   Plot (a') shows a 2D plot of events extracted along the orange edge of plot (a). 
   The V-shaped events near the center are generated by the bouncing ball on the racket, whereas the upper left events are generated by the top of the swinging racket, and the lower right events are generated by the bottom of the swinging racket.
   (b) event images.
   Image (b - i) is captured at $t_{imp} - 6300$ $\mu \mathrm{s}$ (the time at the left edge of plot (a')), whereas image (b - ii) is captured at $t_{imp}$ (the time at the black line of plot (a')).
   (c) PATS images as in Eq.\,\ref{eq:pats_convolution}, corresponding to images (b - i) and (b - ii).
   Image (c - ii) detected more ball events than image (c - i).
   }
   \label{fig:timing_at_impact_2d}
\end{figure}


\begin{figure}[t]
  \centering
   \includegraphics[width=1.0\linewidth]{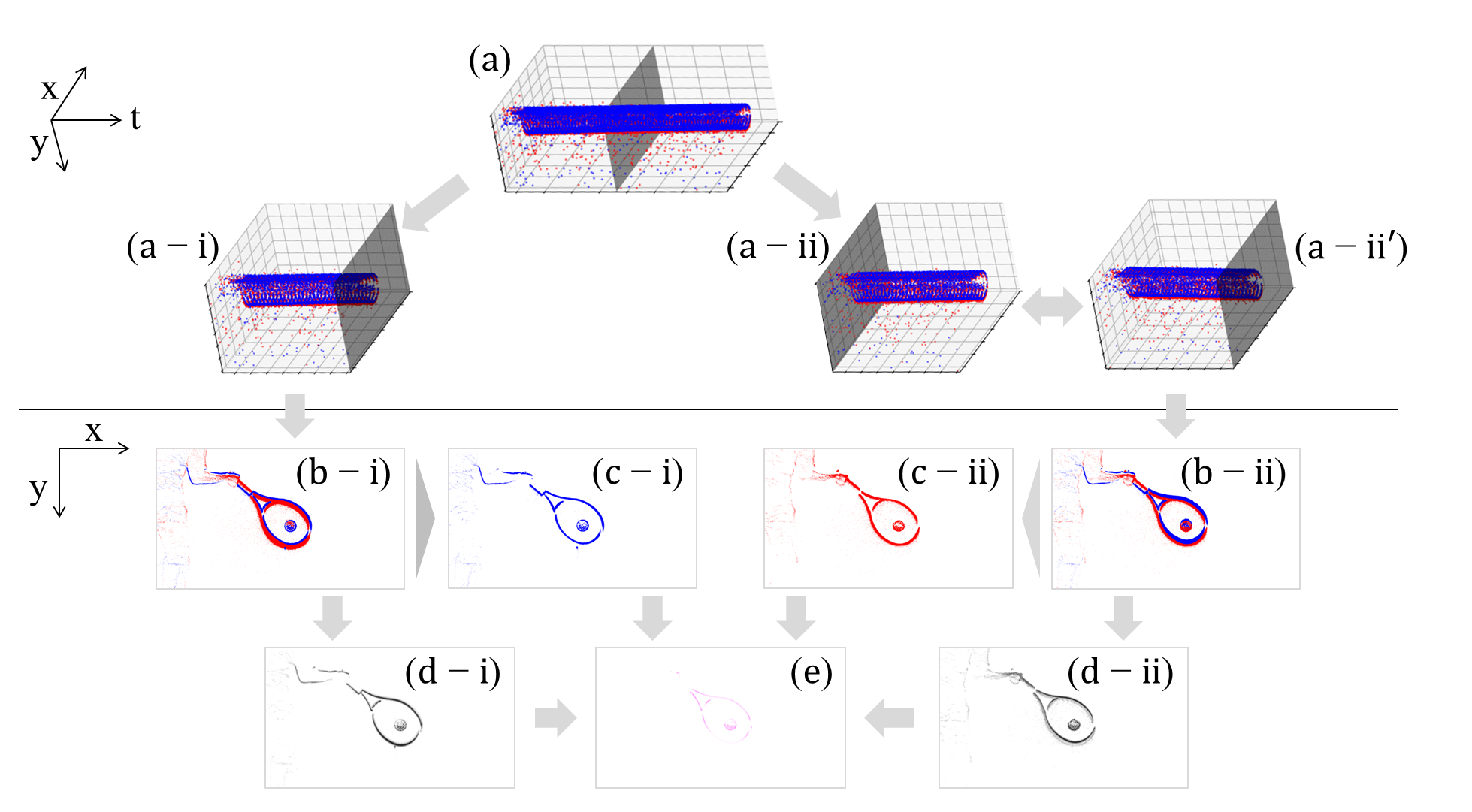}
   \caption{Timing at impact procedure.
   (a) 3D plot of events $e_k$, where blue points represent $+$; red points represent $-$; and the black plane represents reference time $t$.
   Plot (a) represents $\varepsilon_{t}$;
   plot (a - i) represents $\varepsilon_{prev}$; 
   plot (a - ii) represents $\varepsilon_{next}$; 
   and plot (a - ii') represents $\varepsilon_{next}^{\leftarrow}$. 
   (b) (c) event images of i: $\varepsilon_{prev}$ and ii: $\varepsilon_{next}^{\leftarrow}$, respectively, where blue pixels represent $+$; red pixels represent $-$; and white pixels represent $none$.
   Image (b - i) represents $\mathbf{F}_{prev}$;
   image (c - i) represents $\mathbf{F}_{prev}^{+}$;
   image (b - ii) represents $\mathbf{F}_{next}$;
   and image (c - ii) represents $\mathbf{F}_{next}^{-}$.
   (d) grayscale image as a focal time function, where black pixels represent $1.0$, and white pixels represent $0.0$ or $none$.
   Image (d - i) represents $\mathbf{G}_{prev}$, and image (d - ii) represents $\mathbf{G}_{next}$.
   (e) PATS image $| \mathbf{F}_{prev}^{+} * \mathbf{G}_{prev} | * | \mathbf{F}_{next}^{-} * \mathbf{G}_{next} |$, which is a pink grayscale image, where pink pixels represent $1.0$ and white pixels represent $0.0$.
   }
   \label{fig:timing_at_impact}
\end{figure}


\begin{figure}[t]
  \centering
   \includegraphics[width=1.0\linewidth]{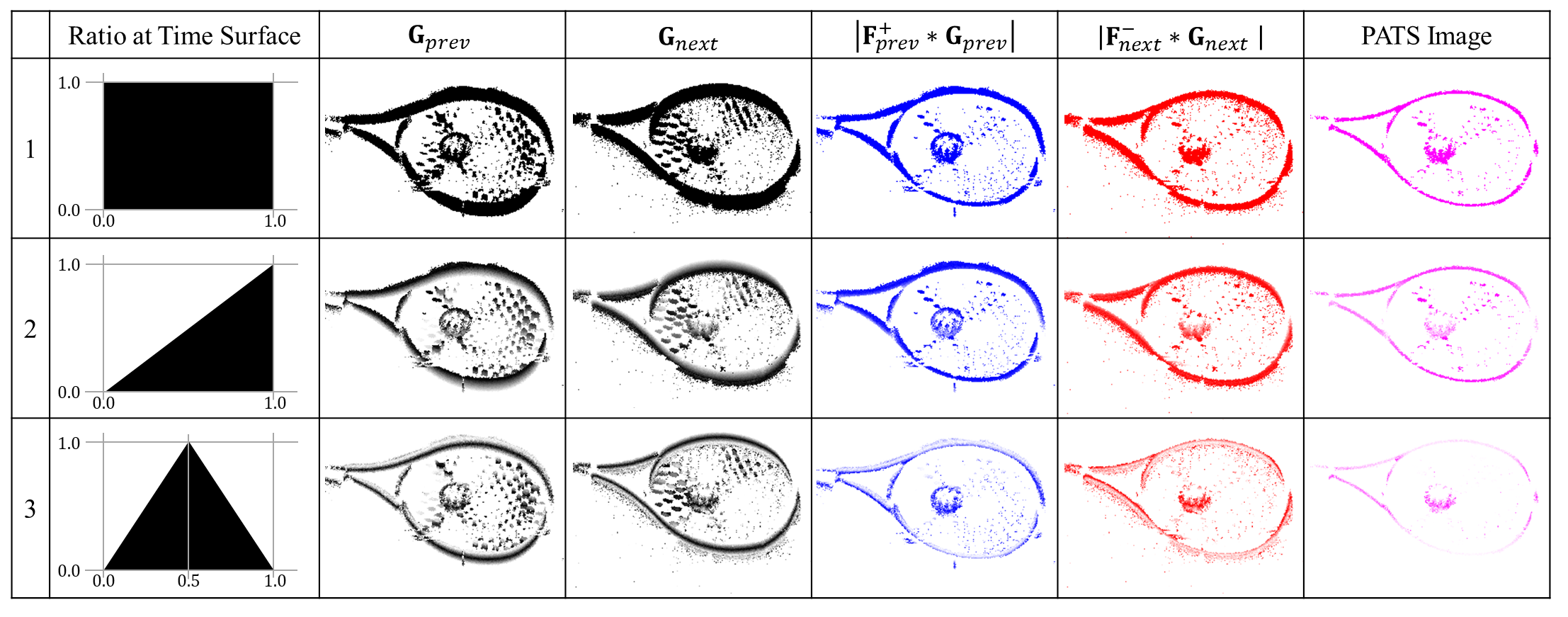}
   \caption{Patterns of various focal time functions.
   Pattern 1 shows a function with uniform ratio ($1.0$), which is the same as an event image. 
   Pattern 2 shows a function where the ratio increases linearly, which is equivalent to a time surface.
   Pattern 3 shows a function where the ratio increases linearly up to $0.5$ time and then decreases linearly up to $1.0$ time.
   The first column is the pattern index.
   The second column shows the ratio at the value of the time surface.
   $\mathbf{G}_{prev}$ and $\mathbf{G}_{next}$ columns show the enlarged grayscale images of the area near the racket, respectively, where black pixels represent $1.0$, and white pixels represent $0.0$ or $none$.
   $| \mathbf{F}_{prev}^{+} * \mathbf{G}_{prev} |$ and $| \mathbf{F}_{next}^{-} * \mathbf{G}_{next} |$ columns show the convolved images, respectively.
   The last column shows the PATS image, which refers to $| \mathbf{F}_{prev}^{+} * \mathbf{G}_{prev} | * | \mathbf{F}_{next}^{-} * \mathbf{G}_{next} |$.
   Pattern 3 most effectively reduces the flickering of the strings.
   }
   \label{fig:convolution}
\end{figure}


\begin{figure}[t]
  \centering
   \includegraphics[width=0.8\linewidth]{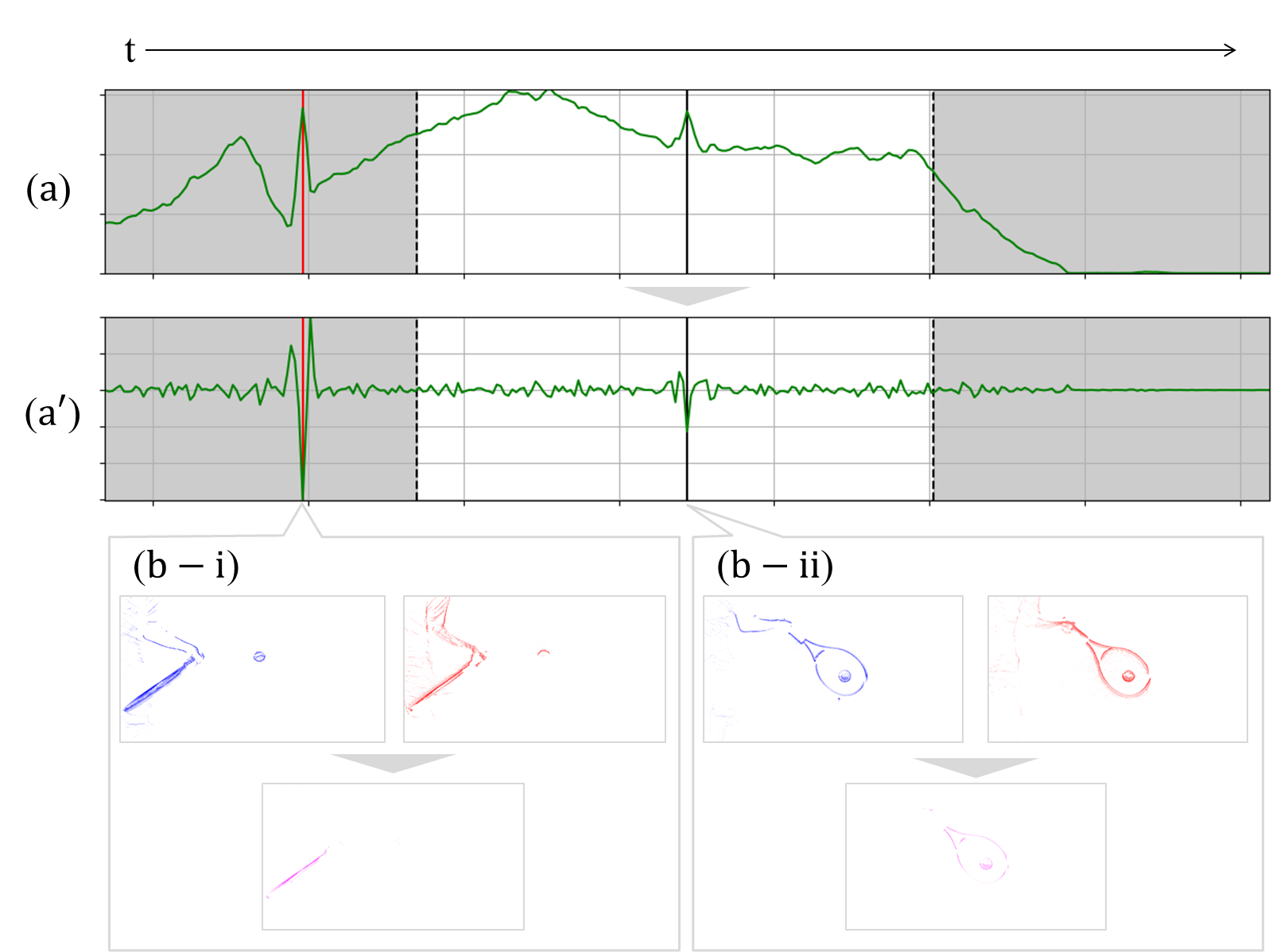}
   \caption{Peaks in the time series of $\rho_{t}$ values (\textit{i.e.\,}PATS).
   Graph (a) time series of $\rho_{t}$.
   Graph (a') shows the Laplacian filtered time series, where the black solid line represents the estimated impact timing $t_{imp}$ within the estimated time range of the swing (black dashed line).
   The red solid line represents false detection if the time range is overestimated.
   (b) results in the PATS image, respectively.
   }
   \label{fig:several_impact_detections}
\end{figure}


\subsection{Timing at Impact}
\label{subsubsec:timingAtImpact}
As shown in Fig.\,\ref{fig:timing_at_impact_2d} (a'), at impact time $t_{imp}$, the ball bounces on the racket.
An event camera captures the bouncing position and shows the changes in the events from  $+$ to $-$ in individual pixels.
We assume that the ball is brighter than the background.

Therefore, we propose an original indicator (PATS: the amount of polarity asymmetry in time symmetry) that focuses on the changes to detect the impact.
As shown in Fig.\,\ref{fig:timing_at_impact}, the procedures for the indicator are as follows:

First, we split the event packet $\varepsilon_{t}$ from reference time $t$ into two packets:
the packet before $t$: $\varepsilon_{prev}$ and the packet after $t$: $\varepsilon_{next}$, defined as
\begin{align}
  \varepsilon_{prev} &= \left\{ {e_k} \in {\varepsilon_{t}} \middle|  {t_k} < {t} \right\} \label{eq:paketPrev},\\
  \varepsilon_{next} &= \left\{ {e_k} \in {\varepsilon_{t}} \middle|  {t} \leq {t_k} \right\} \label{eq:packetNext}.
\end{align}
We reverse $\varepsilon_{next}$ in chronological order to detect whether the events changed from $+$ to $-$ considering the symmetry about $t$.
The reversed $\varepsilon_{next}$ is defined as 
\begin{equation}
  \varepsilon_{next}^{\leftarrow} = \mathrm{reverse}(\varepsilon_{next}) \label{eq:packetNextReversed},
\end{equation}
where $\mathrm{reverse}(\cdot)$ is a function that sorts rows in descending chronological order.

Second, the event images of $\varepsilon_{prev}$ and $\varepsilon_{next}^{\leftarrow}$ are generated, as explained by Prophesee \cite{PropheseeBaseFrame}.
They are defined as
\begin{align}
  \mathbf{F}_{prev} &= \mathrm{image}(\varepsilon_{prev}) \label{eq:imagePrev},\\
  \mathbf{F}_{next} &= \mathrm{image}(\varepsilon_{next}^{\leftarrow}) \label{eq:imageNext},
\end{align}
where $\mathrm{image}(\cdot)$ is a function that generates a $height \times width$ $(\mathrm{px})$ event image and consists of $\{1, -1, 0\}$ assigned as `$+$' to $1$, `$-$' to $-1$, and `$none$' to $0$, respectively.
Note that the pixels of $\mathbf{F}_{next}$ are overwritten by the older events because $\varepsilon_{next}^{\leftarrow}$ is sorted in descending chronological order.

Third, the time surfaces of $\varepsilon_{prev}$ and $\varepsilon_{next}^{\leftarrow}$ are generated.
They are defined as
\begin{align}
  \mathbf{T}_{prev} &= \mathrm{timeSurface}(\varepsilon_{prev}) \label{eq:tsPrev},\\
  \mathbf{T}_{next} &= \mathrm{timeSurface}(\varepsilon_{next}^{\leftarrow}) \label{eq:tsNext},
\end{align}
where $\mathrm{timeSurface}(\cdot)$ is a function that generates a $height \times width$ $(\mathrm{px})$ time surface and  linearly transforms the oldest time to $0.0$ and the most recent time to $1.0$.
Similarly to $\mathbf{F}_{next}$, note that $\varepsilon_{next}^{\leftarrow}$ is sorted in descending chronological order.
The obtained time surfaces are transformed using $\mathrm{focalTime}(\cdot)$, which focuses on the selected time.
They are defined as
\begin{align}
  \mathbf{G}_{prev} &= \mathrm{focalTime}(\mathbf{T}_{prev}) \label{eq:timeInformationPrev},\\
  \mathbf{G}_{next} &= \mathrm{focalTime}(\mathbf{T}_{next}) \label{eq:timeInformationNext}.
\end{align}
The focal time images $\mathbf{G}_{prev}$ and $\mathbf{G}_{next}$ are such that the pixel values of the time surface at the most focused time are set to 1, and the values that are the least focused are set to 0, as shown in Fig.\,\ref{fig:convolution}.

Fourth, we convolve the event images with the focal time images, and calculate the sum of the convolved image.
This is defined as 
\begin{equation}
  \rho_{t} = \sum (| \mathbf{F}_{prev}^{+} * \mathbf{G}_{prev} | * | \mathbf{F}_{next}^{-} * \mathbf{G}_{next} | ), \label{eq:pats_convolution}
\end{equation}
The indicator $\rho_{t}$ is referred to as PATS, and the convolved image is the PATS image.
The reason for convolving the focal time images is to enhance robustness under direct sunlight conditions.
This is because the flickering of the tennis strings due to sunlight causes false impact timing.
Considering that this flicker ($100$ $\mu \mathrm{s}$) takes less time than the impact ($4000$ $\mu \mathrm{s}$, as demonstrated by Cross \cite{cross1999dynamic}), we designed a focal time function to exclude the flicker time, as shown in Fig.\,\ref{fig:convolution}.

Finally, the peak in the time series of $\rho_{t}$ values is detected to identify $t_{imp}$, as shown in Fig.\,\ref{fig:several_impact_detections}.
The Laplacian filter is utilized for peak detection, and the time corresponding to the smallest value in the filtered time series is the estimated impact time $t_{imp}$.

Alternatively, $n_{c}$ candidates for $t_{imp}$ can be obtained in ascending order of peak size, and the one with the centroid position in the PATS image closest to the center can be selected.
This prevents false detection of times when the racket frame overlaps near the start of the swing due to a broad estimation of the swing interval, as shown in Fig.\,\ref{fig:several_impact_detections} (b - i).

In the next step, using $t_{imp}$, contours of the ball and racket are identified at impact.

\subsection{Contours of Ball and Racket}
\label{subsubsec:contoursOfBallAndRacket}
\begin{figure}[t]
  \centering
   \includegraphics[width=0.6\linewidth]{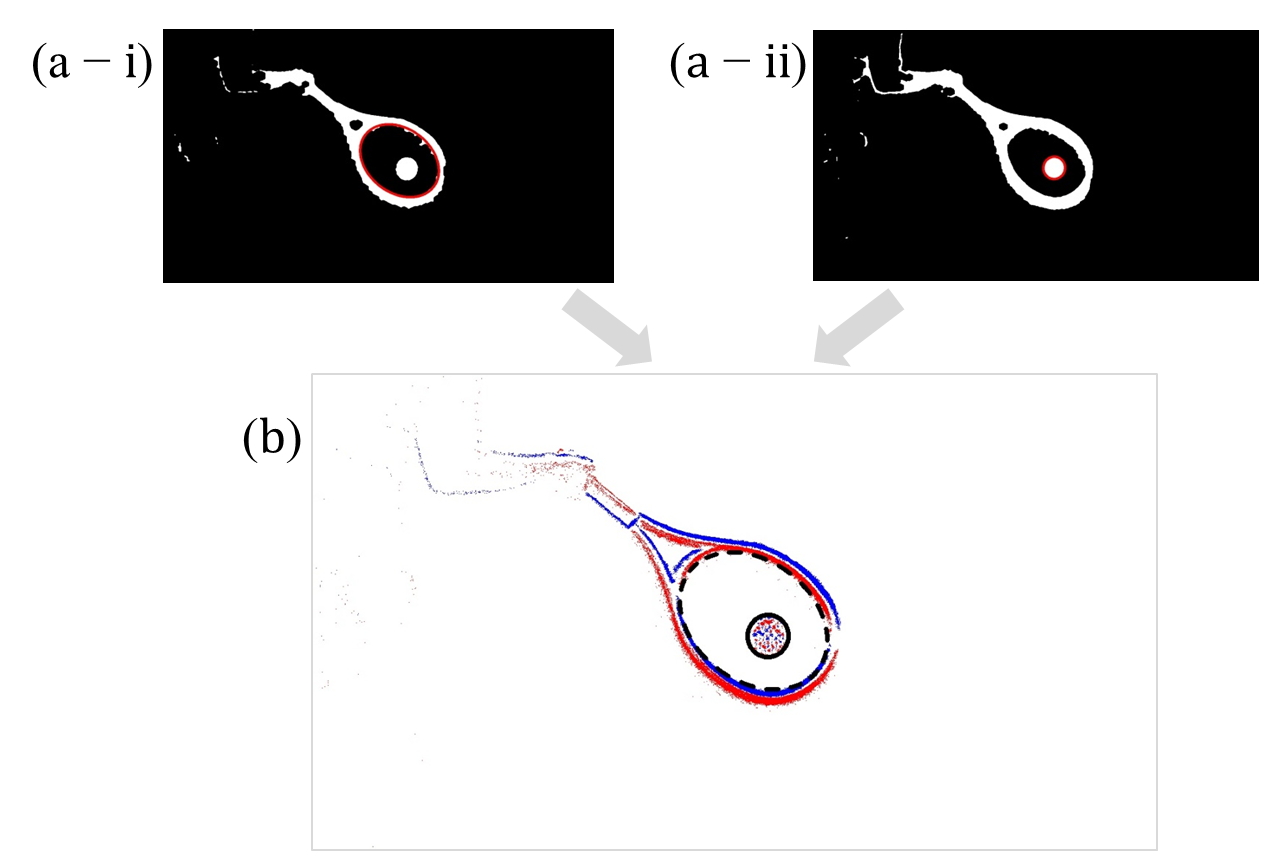}
   \caption{Ball and racket contour procedure.
   (a) binary images at the impact timing $t_{imp}$ with different accumulation times $t_{acc}$.
   Image (a - i) detects the racket with $t_{acc} = 500$, and image (a - ii) detects the ball with $t_{acc} = 2000$, where the red ellipse represents the detected objects.
   (b) event image at $t_{imp}$ with $t_{acc} = 500$ and ellipse detection overlaid, where the black solid line represents the ball's ellipse and the dashed line represents the racket's ellipse.
   }
   \label{fig:contours_of_ball_and_racket}
\end{figure}

Assume that the timing at impact $t_{imp}$ has been accurately obtained, the ball and racket are visually recognized in an event image at $t_{imp}$.
To locate the ball's position on the racket, we identified the approximate contours of the ball and racket as ellipses in this step, as shown in Fig.\,\ref{fig:contours_of_ball_and_racket}.

First, the appropriate accumulation times for the ball and racket are set because the number of events differs depending on the difference in speed as captured by an event camera.
We generated the event images for the ball $\mathbf{F}_{t_{imp}}^{b}$ and racket $\mathbf{F}_{t_{imp}}^{r}$ at $t_{imp}$ using the $\mathrm{image(\cdot)}$ function, as shown in Eqs.\,\eqref{eq:imagePrev} and \eqref{eq:imageNext}.
In the above generation, the activity noise filter was applied to remove noise from these images, as explained by Prophesee \cite{PropheseeActivity}.

Subsequently, $\mathbf{F}_{t_{imp}}^{b}$ and $\mathbf{F}_{t_{imp}}^{r}$ are converted to the binary images $\mathbf{B}_{t_{imp}}^{b}$ and $\mathbf{B}_{t_{imp}}^{r}$, respectively. 
In these binary images, each pixel event is assigned a value of $1$ for $\{+, -\}$ or $0$ otherwise ($none$), as shown in Fig.\,\ref{fig:contours_of_ball_and_racket}.
We applied a morphological transformation that performs a closing operation on these binary images to improve the accuracy of ellipse detection, as explained by OpenCV \cite{OpenCVMorphological}.
To prevent false detections, we can optionally crop the images to the region with a large number of events (\textit{i.e.\,}ROI) in advance.

Finally, ellipse detection is applied to $\mathbf{B}_{t_{imp}}^{b}$ and $\mathbf{B}_{t_{imp}}^{r}$, as explained by OpenCV \cite{OpenCVEllipse}.
This detection retrieves the properties of the ellipses, which include the center coordinates, semi-major axis, semi-minor axis, and rotation angle.
In $\mathbf{B}_{t_{imp}}^{r}$, the second largest detected ellipse is estimated to be inside the racket frame, whereas in $\mathbf{B}_{t_{imp}}^{b}$, the largest detected ellipse inside the frame is estimated to be the ball.

\subsection{Output}
\label{subsec:output}
As output, the relative location of the ball on the racket is expressed as a percentage.

The center of the ball relative to the racket is located using the properties of the ellipses.
In particular, the center coordinates of the ball's ellipse are mapped to the axes of the racket's ellipse in the $uv$ coordinate system, and the ratios of the axes are calculated.
Note that the $+u$ direction is defined as upward along the racket, and the $+v$ direction is defined as toward the tip of the racket.

\begin{figure}[t]
  \centering
   \includegraphics[width=0.45\linewidth]{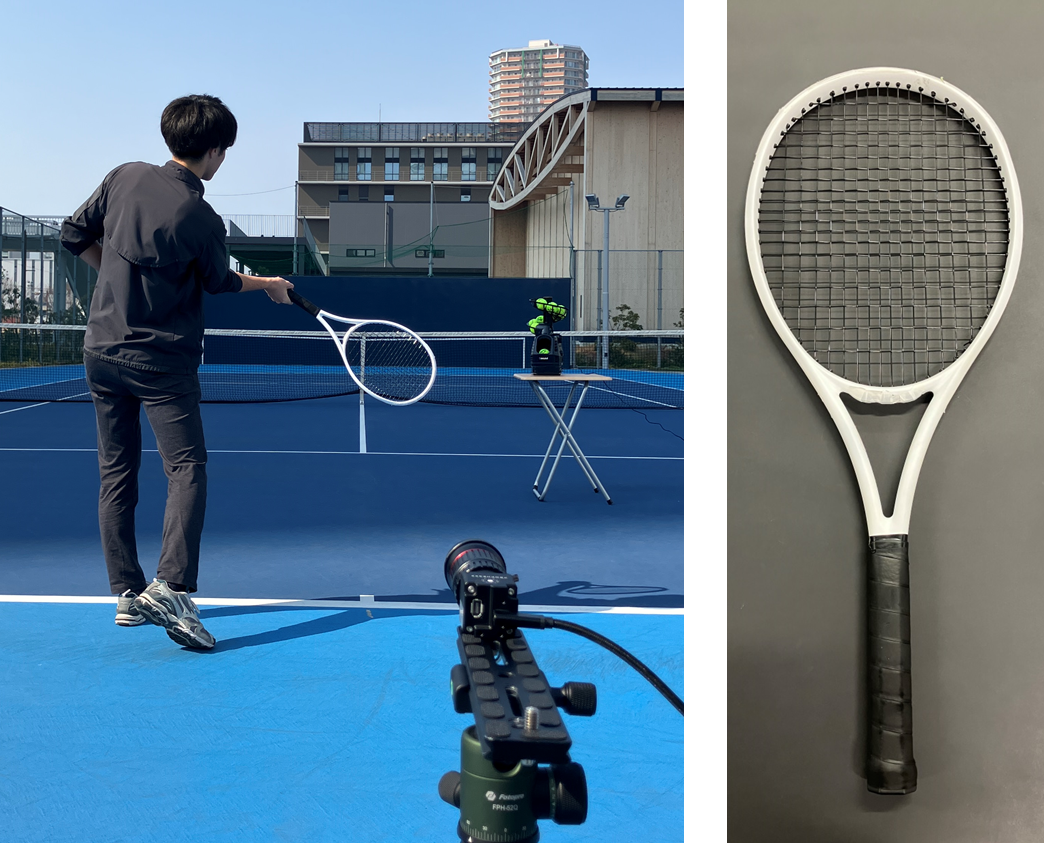}
   \caption{Play scene and equipment for our experiments.
   The event camera captured the scene from behind the player, and the ball was tossed from the machine located in the front right, as shown in the left figure. 
   In the case of left-handed players, the position of the machine was reversed.
   A white racket frame and black polyester strings were used, as shown in the right figure.
   }
   \label{fig:setup}
\end{figure}


\section{Experiment}
\label{sec:experiment}
To verify the proposed method, we conducted experiments to capture tennis scenes using an event camera.

\subsection{Setup}
\label{subsec:setup}
For the event camera, SilkyEvCam HD (CenturyArks \cite{CenturyArks25}) was used for these experiments.
The laptop used for processing event data was equipped with an Intel Core i7-12700H CPU (14 cores, 20 threads, 2.3 GHz), an NVIDIA GeForce RTX 3080 Ti Laptop GPU (16 GB GDDR6 memory), and 32 GB of LPDDR5 memory (5200 MHz).
This method was implemented in Python.

\subsection{Play Scene}
\label{subsec:playScene}
As shown in Fig.\,\ref{fig:setup}, the ball impact was measured on an outdoor tennis court.
The camera was positioned at a height of $0.7$ $\mathrm{m}$ to match the impact location and then placed $2.7$ $\mathrm{m}$ behind it.
A zoom lens was used for the measurement. 
The data were captured both under direct sunlight conditions and in the absence of sunlight.

These tennis scenes were designed as rallies, where a player consecutively hit back 12 balls from a tossing machine, and then each ball bounced once after being tossed.

Three players were right-handed, whereas two players were left-handed.
The equipment consisted of a white racket and black polyester strings to improve the accuracy of measuring the impact location by capturing changes in light more distinctly.
Specifically, this setup enhanced the events for the ball and racket frame while reducing the events for the strings.

\subsection{Metrics}
The visual definitions of the time and position at impact using an event camera are based on those of Yasuda \textit{et~al.\,}\cite{Yasuda24}.
The impact timing is defined as the moment when the area of the ball for $+$ events is at its maximum.
We visually inspected the timing and location at impact, using the following definitions.

Each of the three identification steps was identified, as follows.
\def\theenumi{\Roman{enumi}}
\def\labelenumi{\theenumi.}
\begin{enumerate}
    \item \textbf{Time Range of Swing:} We verified whether the visually inspected timing was included in the identified swing section.
    If it was included, the next step was proceeded to as a success; if it was not included, the next step was not proceeded to, considering it a failure.
    \item \textbf{Timing at Impact:} The absolute difference in time ($\mu\mathrm{s}$) between the identified and visually inspected timing was verified.
    If the difference was 2000~$\mu\mathrm{s}$ or less, the next step was proceeded to as a success; if it was greater, the next step was not proceeded to, considering it a failure.
    Note that Cross \cite{cross1999dynamic} demonstrated that the contact time of the ball during the impact in tennis is approximately $4000$ $\mu \mathrm{s}$.
    \item \textbf{Contours of Ball and Racket (Output):} The relative absolute difference in percentage points ($\%\mathrm{pt}$) between the identified and visually inspected location was verified. 
    If the contours were not detected, the step was considered a failure.
    Note that this location was utilized because it is easier to evaluate than the contour.
\end{enumerate}


\begin{table}
  \caption{Variables of our proposed method were set using the training data.}
  \label{tab:variables}
  \centering
  \begin{tabular*}{70mm}{@{\extracolsep{\fill}}ccr}
    \toprule
    Section & Symbol & Value\\ 
    \midrule
    \multirow{6}{*}{Time Range of Swing} & $t_{acc}$ & $500$\\
                                         & $t_{strd}$ & $500$\\
                                         & $n_{\varepsilon}$ & $10$ \\
                                         & $\tau_{mean}$ & $1 \times 10^{7}$ \\
                                         & $\tau_{var}$ & $6 \times 10^{11}$ \\
                                         & $\tau_{t}$ & $100000$ \\ \hline
    \multirow{3}{*}{Timing at Impact}    & $t_{acc}$ & $4000$\\
                                         & $t_{strd}$ & $500$\\
                                         & $n_{c}$ & $3$\\ \hline
    Contours of Ball & $t_{acc}$ & $2000$\\ \hline
    Contours of Racket & $t_{acc}$ & $500$\\
    \bottomrule
  \end{tabular*}
\end{table}

\subsection{Variables}
For the 60 data obtained from 12 consecutive balls hit by 5 players, the first 2 balls were used as training data (a total of 10 data), whereas the subsequent 10 balls were used as test data (a total of 50 data).
Using the training data, the variables were set and summarized in Table \ref{tab:variables}.
For the focal time function, we selected Pattern 3 in Fig.\,\ref{fig:convolution}.

\subsection{Result}
\label{subsec:result}
Verification was conducted using 10 test data from each of the 5 players, totaling 50 data (instances).
Players No. 1 -- 3 were measured in the absence of sunlight, whereas No. 4 and 5 were measured under direct sunlight conditions. 
Additionally, No. 3 and 5 are left-handed.
The results in Table \ref{tab:result} are explained as follows. The mean computation time per instance was less than 2 seconds.

I.\, \textbf{Time Range of Swing:}
A total of 49 out of 50 instances were correctly estimated.
Specifically, for player No.1, one instance of range estimation did not include the visually inspected impact timing. 
For players No.4 and 5, one instance each of overestimating the swing range occurred.

II.\, \textbf{Timing at Impact:}
A total of 46 out of 49 instances were correctly detected. 
Specifically, the three instances that were not correctly detected involved misestimations of the timing: one instance was estimated to be 26900 $\mu \mathrm{s}$ (No.2) after the visually inspected impact timing, and two instances were estimated to be 8300 $\mu \mathrm{s}$ (No.2) and 13500 $\mu \mathrm{s}$ (No.3) before it.

III.\, \textbf{Contours of Ball and Racket (Output):}
In the absence of direct sunlight, 24 out of 26 instances were successfully detected for contour detection.
Whereas under direct sunlight, only 3 out of 20 instances were detected.
For all instances where contour detection was successful, the relative absolute difference was less than $12.1$~$\%\mathrm{pt}$ for $u$ and $9.1$~$\%\mathrm{pt}$ for $v$.
This means the difference was kept below $15$~$\mathrm{mm}$, corresponding to less than one-quarter of the diameter of a tennis ball.


\begin{table}[t]
  \centering
  \caption{Results of our experiments.
  The ratio at each step represents the number of successful instances / total number of instances.
  For players No. 4 and 5, the time range of the swing was overestimated by one instance each.
  Output indicates that the red plots represent the visually inspected impact positions. 
  The crosses ($\times$) represent instances that failed in the time range of swing or timing at impact, while the triangles ($\triangle$) indicate failures in the contours of the ball and racket.
  The circles ($\bigcirc$) indicate successful estimation instances, with black circles representing the estimated impact positions, and their diameter corresponds to one-quarter the size of a tennis ball.
  The black solid lines connect the corresponding visually inspected and estimated impact positions. 
  }
  \label{tab:result}
  \makebox[1 \textwidth][c]{
  \resizebox{1.0 \textwidth}{!}{
  \begin{tabular}{c|ccccc}
    \hline
    Player No. & 1 & 2 &  3 & 4 & 5
    \\ 

    Under Direct Sunlight & & & & $\surd$ & $\surd$
    \\

    Left-Handed & & & $\surd$ &  & $\surd$ 
    \\

    \hline
    Time Range of Swing & $9 / 10$ & $10 / 10$ & $10 / 10$ & $11 / 10$ & $11 / 10$
    \\

    \hline
    Timing at Impact ($\mu\mathrm{s}$) & 
    \begin{tabular}{c} $322 \pm 244$ \\ $9 / 9$\end{tabular} & 
    \begin{tabular}{c} $350 \pm 240$ \\ $8 / 10$\end{tabular} &
    \begin{tabular}{c} $322 \pm 79$ \\ $9 / 10$\end{tabular} &
    \begin{tabular}{c} $460 \pm 254$ \\ $10 / 10$\end{tabular} &
    \begin{tabular}{c} $430 \pm 224$ \\ $10 / 10$\end{tabular}
    \\

    \hline
    Output ($\%$) &
    \begin{tabular}{c} 
    \begin{minipage}{2truecm}
      \centering
      \includegraphics[width=2truecm,clip]{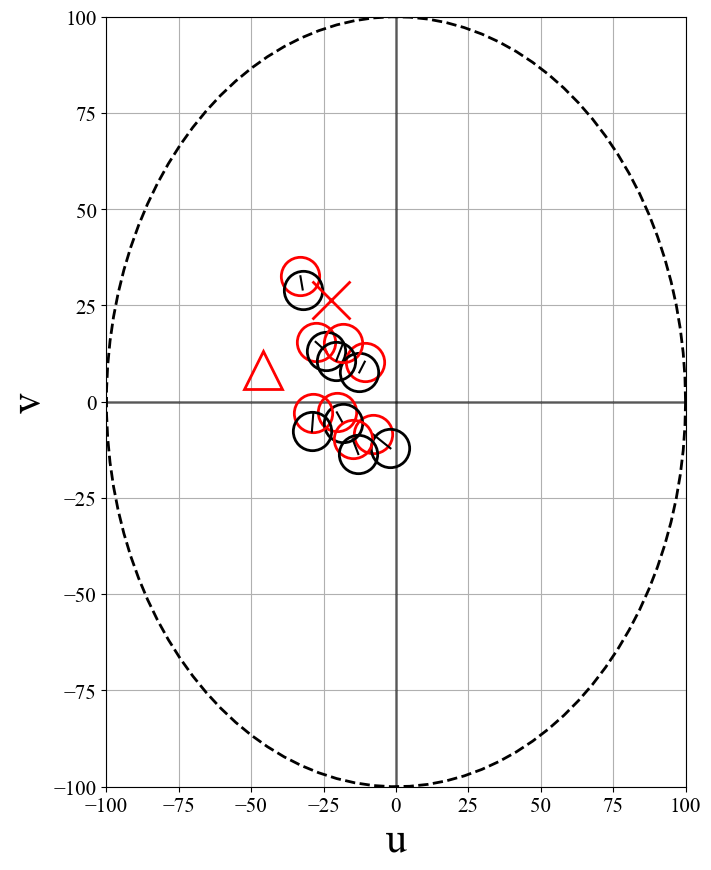}
    \end{minipage}
    \\ $8 / 9$
    \end{tabular}
    &
    \begin{tabular}{c} 
    \begin{minipage}{2truecm}
      \centering
      \includegraphics[width=2truecm,clip]{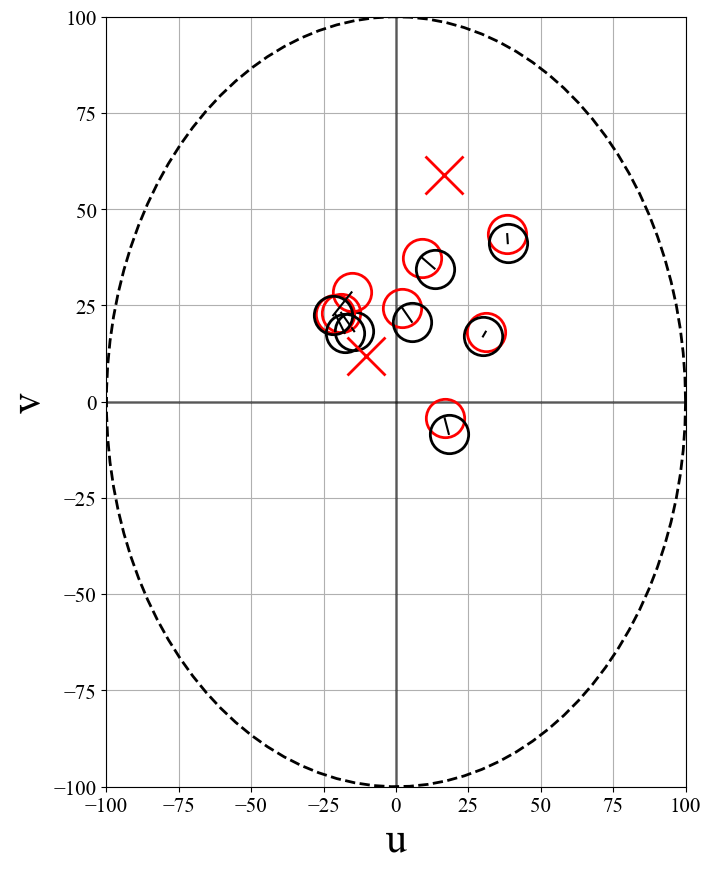}
    \end{minipage}
    \\ $8 / 8$
    \end{tabular}
    &
    \begin{tabular}{c} 
    \begin{minipage}{2truecm}
      \centering
      \includegraphics[width=2truecm,clip]{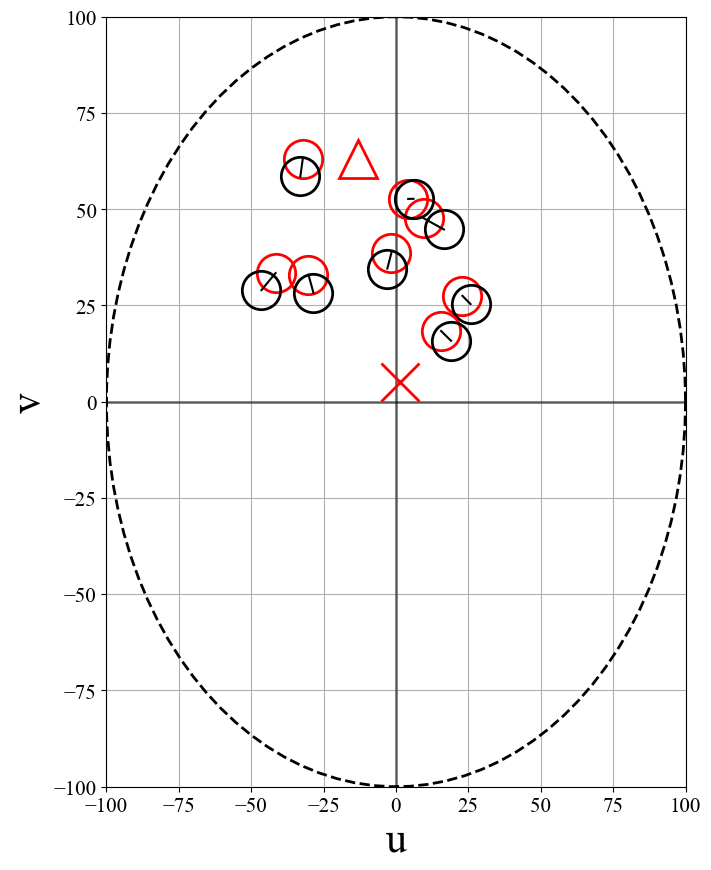}
    \end{minipage}
    \\ $8 / 9$
    \end{tabular}
    &
    \begin{tabular}{c} 
    \begin{minipage}{2truecm}
      \centering
      \includegraphics[width=2truecm,clip]{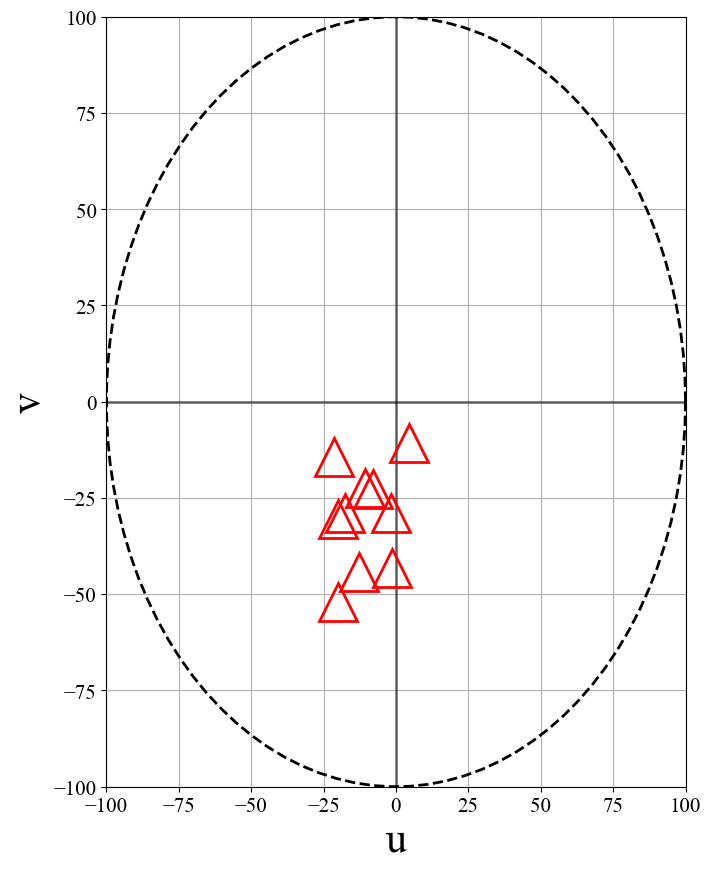}
    \end{minipage}
    \\ $0 / 10$
    \end{tabular}
    &
    \begin{tabular}{c} 
    \begin{minipage}{2truecm}
      \centering
      \includegraphics[width=2truecm,clip]{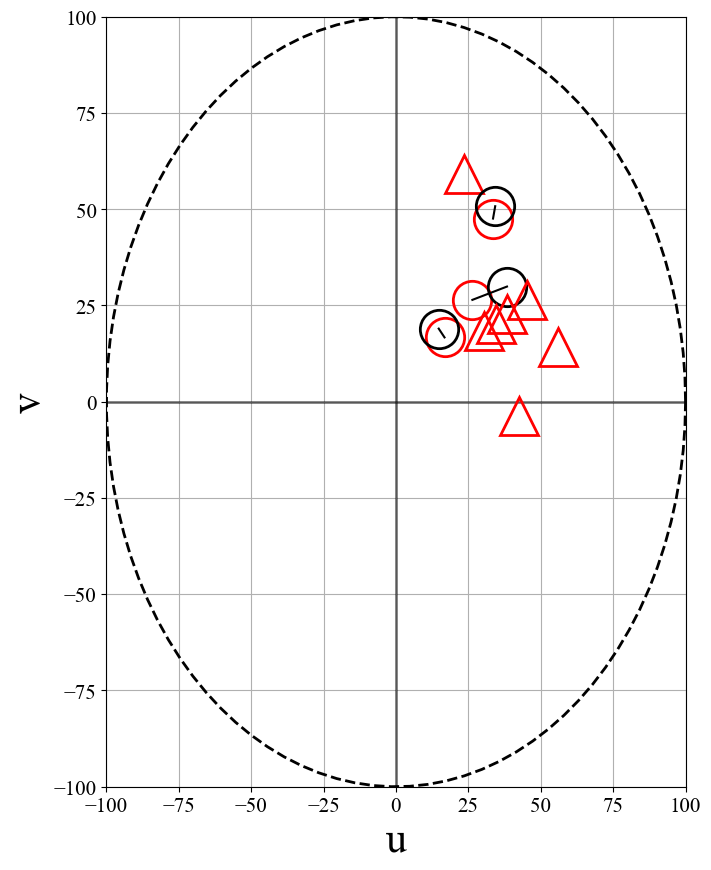}
    \end{minipage}
    \\ $3 / 10$
    \end{tabular}
    \\

    \hline
    Computation Time ($\mathrm{s}$) & $1.83 \pm 0.34$ & $1.86 \pm 0.25$ & $1.41 \pm 0.11$ & $1.75 \pm 0.23$ & $1.97 \pm 0.19$
    \\

    \hline
  \end{tabular}
  }
  }
\end{table}
\section{Discussion}
\label{sec:discussion}

\subsection{Consideration}
\label{subsubsec:consideration}
Out of a total of 50 test data (instances), the impact timing was correctly estimated for 46 instances.
Our proposed PATS was effective even under direct sunlight conditions.
Thus, it is suggested that this method eliminates the effort of manually searching for the impact timing.

For all instances where contour detection was successful, the relative absolute difference in the impact position was within the permissible range for measuring tennis players' performance as indicated by Yasuda \textit{et~al.\,}\cite{Yasuda24}.
This range corresponds to less than one-quarter of the diameter of a tennis ball ($15$ $\mathrm{mm}$). 
We confirmed that the impact positions are distributed at various locations for each player.
Therefore, it is suggested that the impact position is useful for analyzing players' and equipment characteristics.

The computation time was generally less than $2$ seconds, indicating that it is shorter than the time taken for a rally.
Thus, our method can be considered a valuable real-time system for analysis.
\subsection{Limitation}
\label{subsubsec:limitation}

\subsubsection{Influence of Direct Sunlight on Detection Rates:}
\begin{figure}[t]
  \centering
   \includegraphics[width=0.7\linewidth]{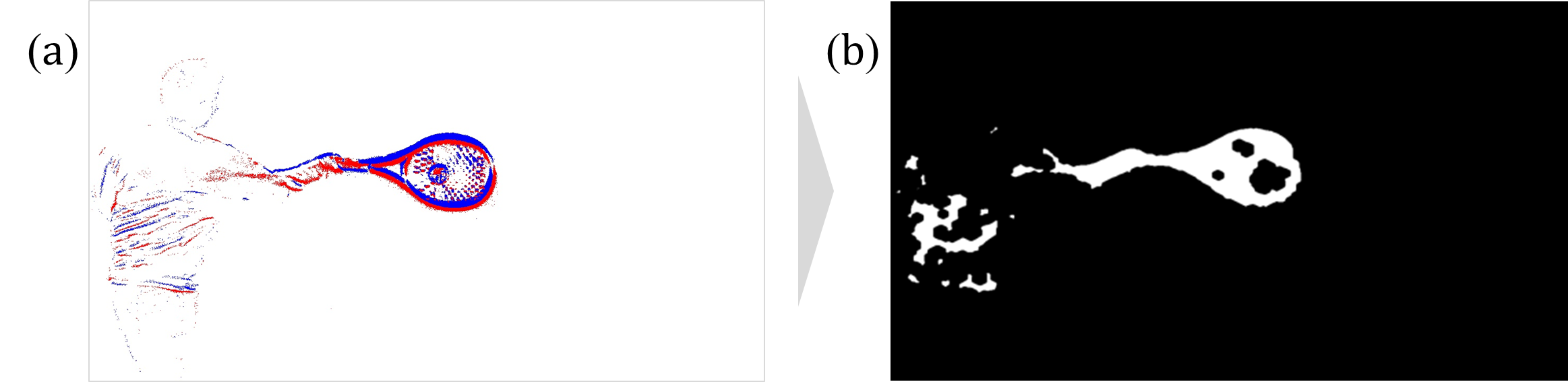}
   \caption{Failure of contour estimation due to string flickering under direct sunlight.
   (a) event image at the impact timing.
   (b) binary image converted from image (a). The flickering of the strings makes it difficult to identify the contour of the ball.
   }
   \label{fig:limitation_reflection_of_sunlight}
\end{figure}

Under direct sunlight conditions, the detection rate for estimating the impact timing did not decrease; however, the detection rate for contour estimation significantly decreased.
As shown in Fig.\,\ref{fig:limitation_reflection_of_sunlight}, this is due to the simultaneous occurrence of events related to both string flickering caused by sunlight reflection and the ball, which makes it difficult to detect the contours of the ball in the binary image.

\subsubsection{Solid-Colored Constraints in Equipment:}
A solid-colored racket and strings were used to clarify the contours of both the racket and the ball. 
In addition, the racket was colored white to create a contrast with the background, whereas the strings were colored black to contrast with the ball, thereby increasing the number of events. 
Due to the characteristics of event cameras, using appropriate solid-colored equipment depending on the capturing environment is necessary. 
Additionally, using equipment with multiple colors may result in event loss at the boundaries of the colors, complicating contour detection.

\subsection{Future Work}
\label{subsubsec:futurework}
To improve the detection rate of contours under direct sunlight conditions, the utilization of the proposed PATS image is considered. 
Additionally, implementing existing motion compensation methods to clarify the contours in the event images may also be a viable approach, including those described in \cite{gallego18,shiba22,stoffregen19}.

Extensions to 3D measurements for estimating the tilt of the racket at impact and adaptations to other sports are also considered.

\section{Conclusion}
\label{sec:conclusion}
We propose a method for locating the tennis ball impact on the racket in real time using an event camera. 
The process comprises three identification steps: the timing range of the swing, the timing at impact, and detecting the contours of the ball and racket. 
Particularly, the impact timing estimation method PATS is suggested to eliminate the effort of manually searching for the impact timing.

The results of the experiments were within the permissible range for measuring tennis players' performance.
Moreover, the computation time was sufficiently short for real-time applications.
Thus, the proposed method can be considered a valuable real-time system for analysis.

\bibliographystyle{splncs04}
\bibliography{main}
\end{document}